\documentclass{article}

\usepackage{arxiv}

\usepackage[utf8]{inputenc} 
\usepackage[T1]{fontenc}    
\usepackage{hyperref}       
\usepackage{url}            
\usepackage{booktabs}       
\usepackage{amsfonts}       
\usepackage{bm}
\usepackage{nicefrac}       
\usepackage{microtype}      
\usepackage{lipsum}
\usepackage{graphicx,wrapfig}
\usepackage{caption,subcaption}
\usepackage[ruled,vlined]{algorithm2e}
\usepackage{float}
\usepackage{enumitem}
\usepackage[english]{babel}
\usepackage[sort, numbers, authoryear]{natbib}
\graphicspath{ {./images/} }

\title{HindSight: A Graph-Based Vision Model Architecture For Representing Part-Whole Hierarchies}

\author{
 Muhammad AbdurRafae \\
  \texttt{m.abdur.rafae@gmail.com} \\
}

\begin{document}
    \begin{center}
        \includegraphics[width=0.2\linewidth]{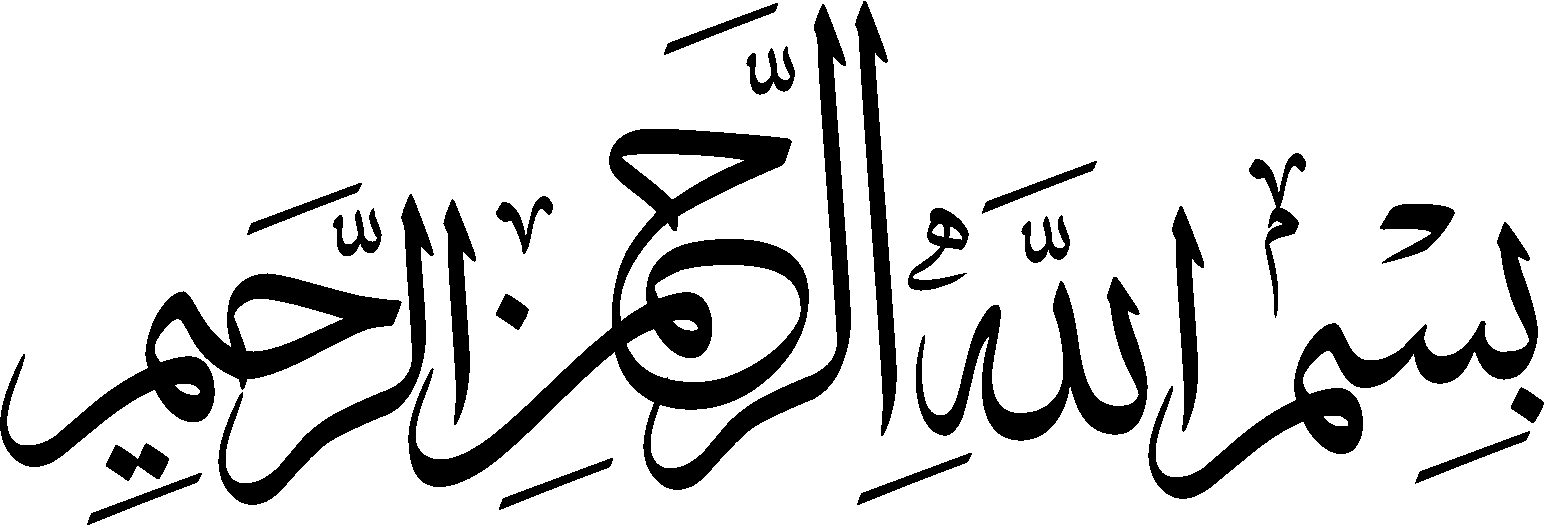}
        \vspace{86pt}
    \end{center}
\maketitle

\begin{abstract}
This paper presents a model architecture for encoding the representations of part-whole hierarchies in images in form of a graph. The idea is to divide the image into patches of different levels and then treat all of these patches as nodes for a fully connected graph. A dynamic feature extraction module is used to extract feature representations from these patches in each graph iteration. This enables us to learn a rich graph representation of the image that encompasses the inherent part-whole hierarchical information. Utilizing proper self-supervised training techniques, such a model can be trained as a general purpose vision encoder model which can then be used for various vision related downstream tasks (e.g., Image Classification, Object Detection, Image Captioning, etc.). 
\end{abstract}

\section{Introduction}
Computer Vision has advanced tremendously in the last few decades. In order to achieve this, we have been using larger datasets, formulating novel algorithms and techniques, and developing optimized and more powerful hardware. However, there are still some issues that cause our current vision models to not be at par with human level. Some of these are elaborated below:

\paragraph{Part-Whole Hierarchies} Our current model architectures neither fully utilize part-whole hierarchies of objects nor the spatial relationship among them. The concept of part-whole hierarchies can be easily explained by taking an example of a human face. Here the "face" represents the whole object and "eyes, ears, nose and mouth" are its parts. Another example is of a car, the model would see a car tyre in most images that contain a car, however it will not implicitly link the two objects (cars and car-tyres) using a hierarchical relation. If a model is able to learn a representation of the \textit{whole} along with representation of its \textit{parts} and the \textit{spatial relationship} between them, then the model is capable of learning a much richer representation from the image. The main difficulty in implementing this type of model is that it requires a tree-like graph to be constructed from different regions (patches) of image in a dynamic fashion. The tree could have different number of levels based on the contents of the image. Face (\textit{part}) could belong to head (\textit{whole}) and head (\textit{part}) could belong to human body (\textit{whole}). Similarly, we can extend the levels on other side, e.g., iris texture (\textit{part}) could belong to eye (\textit{whole}). 

\paragraph{Lack of Scale In-Variance}  Another major issue is the lack of scale in-variance in our models, although data augmentation techniques and usage of pooling layers have helped to a small degree. Addition of residual connections in CNNs also provide a little robustness to scale in-variance by allowing higher layers to have access to low level raw features via the skip connections. However, objects and their parts usually have very different scales in images, which worsens the problem of scale in-variance for the model. Assuming we create a huge labelled dataset where we have images of objects as well as for the parts that make up these objects. For example, our dataset would contain (among other classes) a class for cars and another class for car\_tyres. Since the same pattern in an image at different scales activates different convolutional filters in different layers, hence if we were to train a model on this dataset, the model would learn two different set of filters to detect the same object (i.e., a car\_tyre). One set (\textit{A}) would be in the deepest layer for classifying car\_tyres and the other set (\textit{B}) would be at a layer of intermediate depth (and would be helping in identification of cars). This is because the \textit{set B} filters in the intermediate layer lack the receptive field required to identify a car\_tyre that is covering the whole image, while the \textit{set A} filters in the deepest layer would be unable to communicate with the filters for identifying cars in images. Hence, lack of scale in-variance in these models causes an in-efficient use of compute resources. For a truly generalized object identification and classification model, this is also an in-efficient use of data since "car tyres" present in training images of the class car would not be implicitly utilized for training for the class of car\_tyres (without manual data manipulation and labelling). 

\paragraph{Fixed Low Input Dimensions} Our current models are bound to a fixed input dimension and usually this dimension is kept to a low value like 224x224 ($\sim$50k Pixels). Any image of higher resolution has to be down-scaled before being processed by the model and this down-scaling causes a loss in valuable information and fine details. We have improved our image capturing technologies to a far greater extent than what we can process with our current vision models. Any common mobile device can take an image of multiple Megapixels while our state of the art models are restricted to only work on images of few Kilopixels or even lower. Although for simple image classification tasks this small size has not been that much of an issue, but as we approach more advanced vision tasks (like image captioning, visual question answering etc) we would need to transition into models that can work on images with larger dimensions and hence extract more information from images. CNNs are the most efficient information extraction models for images of low input dimensions. However, the rigidity in the structural inductive priors of convolution kernels responsible for this high efficiency in low dimensions hinders them to perform well when the image size is increased. 

\medskip

We need to address these issues in order to create a generalized vision model. HindSight attempts to provide a solution for these issues (\textit{part-whole hierarchies}, \textit{lack of scale in-variance} and \textit{fixed low input dimensions}).

\section{Background}

\cite{AlexNet} (AlexNet) displayed the importance and capability of CNNs in vision tasks. This model vastly outperformed traditional vision models on the ImageNet dataset. This was essentially a turning point for research in the field of vision models. \citet{ResNet} (ResNet) demonstrated that by using residual connections, a model could be trained with much more depth. These deeper models provided a boost in performance when the depth was increased to moderate levels (e.g., 50, 100, 150) however increasing the depth further (e.g., 1000+) provided very diminished returns. Since then, researchers have found novel and optimized methods of scaling up the CNNs to improve performance while utilizing similar blocks of model architecture.

\cite{AttentionPaper} (Attention is all you need) introduced the transformer architecture which is built using attention mechanism. The transformer first creates a graph using all of the input tokens (along with positional encoding) as nodes and then uses the attention masks to define the adjacency matrix for these nodes. The model then encodes the provided inputs into a generalized graph representation, which can then be utilized for downstream tasks. Initially transformers were introduced for language translation, however they have been demonstrated to perform comparable to state of the art in various other domains with minimal modifications in the original architecture. \cite{VitPaper} (Vision Transformer), applied a pure transformer directly to a sequence of image patches and attained results comparable to state of the art CNNs. In this architecture, the images are broken up into small 16x16 dimension patches and then converted into a graph of patches for the transformer blocks to process. The refined graph representation, obtained after the transformer blocks, is then utilized to classify images.

\cite{part_whole} (Neural Models for Part-Whole Hierarchies) presented a method for representing images exploiting their inherent hierarchical nature. They explored both bottom-up and top-down connections for representing information and provided a working example using cartoon faces. Recently, \cite{CapsuleNet} (Capsules) and \citet{glompaper} (GLOM) introduced new concepts and architectures for learning part-whole hierarchies of objects in images. They propose architectures to the mimic multi-layer visual system of humans, which creates a parse tree-like structure to understand the provided visual input. In \cite{CapsuleNet} the main component of the architecture is a capsule, all possible objects and parts have their own separate capsules. A higher level capsule represents a whole (e.g., a face) and is activated by lower level capsules of parts (e.g., nose, eyes, mouth). A consequence of this specific setup of independent capsules is that parts of different objects which look similar can not be learned by the same capsule. This issue can be rectified by substituting these part specific capsules with universal capsules. \citet{glompaper} addresses this issue and proposes a new architecture, where a multitude of such universal capsules are applied to each location in the image. Each set of these universal capsules capture a different level of information from the image. The result is a multi-level representation of each location of the image. The information between the levels and adjacent locations is then propagated until they converge to a stable state. The idea is that by doing this propagation of information till convergence, the representations would form clusters where the higher level representations would converge in patches for each of the distinct objects and the lower level representations would converge for patches that belong to same part in an object. GLOM represented the idea in form of an imaginary architecture for vision models where the model would be able to learn the part-whole hierarchies from the image. HindSight is an humble attempt at a working model architecture to reach the same end goal.

\section{Basic Intuition}

The main goal of HindSight is to create a model with the capacity to represent images in form of a graph which encompasses part-whole hierarchies and the structural relationships (both inter-objects and intra-object) present in the images. 

In order to learn the part-whole hierarchies, first the individual representations of objects (whole) and their parts have to be learned by a model. The issue in learning their representations together is that there is a large difference in scale between the objects and their parts. The model should be able to recognize objects in multiple scales since the same object can be of different scale in different images. For example, a car tyre in \textit{(A)} a highway full of cars, \textit{(B)} an individual car and \textit{(C)} a close-up of single car tyre. Since almost all of our current state of the art model architectures are neither scale invariant nor do they capture the inherent structural relationship between objects and parts, hence the only method available to learn individual representation of objects as well as individual representation of their parts is to train a model on a huge labelled dataset comprising of objects and their parts. This would enable us to train a model that can work on multiple scales without worrying about which scale the given input image belongs too. A single model can be constructed to achieve this. Since the variation in the types of inputs at different scales is too large, hence this single model would have to be quite wide and deep to be able to learn the appropriate representations. In the last decade, a bulk of researched and implemented Computer Vision models have followed the same suit of increasing depth and width of model to deal with this issue.

As an alternative, we can create a set consisting of multiple models, where each model would only learn to represent objects of one specific scale. The issue now would be to determine which model should work on which scale of objects. We'll discuss the resolution of this issues later in this section and carry on assuming that this has been resolved. This set of models enables us to to extract individual representation from images of objects as well as from images of their parts.

However, this learned set of models is still not able to determine the inherent structural relationship among the objects and their parts. In order to extract the structural relationship between them, we need a model with the ability to parse the image into a tree-like structure and learn the hierarchical relationship between objects and their parts. This means that we need to create a graph representation of the image.

Some recent papers have used the concept where the image is broken up into a grid of small patches and each patch of the image represents a node in a fully connected graph. After a sufficient number of graph iterations, the updated node representations of the graph are used for downstream vision tasks (like Image Classification). This approach can be split into two modules, where the first module (Feature Extractor) only learns to represent the parts of an object and the second module (Attention Blocks) learns to represent the whole image from the representations of individual parts/patches after multiple iterations for updating these node representations. Although this approach does learn a graph representation of the image, it only incorporates the relationship of one parts of an object to another part. The relationship of a part of an object to the whole object can not be learned since there is no node present in the graph that can represent this object as a whole. 

One simple way to resolve this issue is to create multiple levels of patches from the same image and use our set of models on each of them. This would enable us create nodes for objects as well as their parts. We can then use a graph neural network to refine these node representations and learn the inherent structural relationship between objects and their parts.

\begin{figure}[t]
    \begin{subfigure}{0.33\textwidth}
          \includegraphics[width=\linewidth]{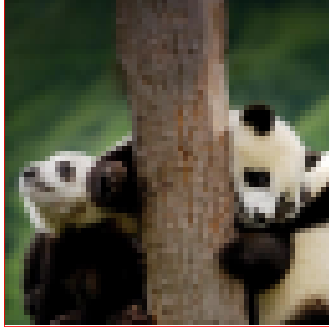}
          \caption{Highest Level Patch}
          \label{highest_panda_image}
    \end{subfigure}\hfil 
    \begin{subfigure}{0.33\textwidth}
          \includegraphics[width=\linewidth]{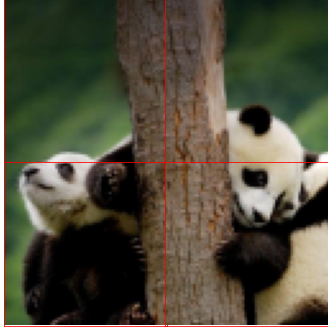}
          \caption{$2^{nd}$ Level Patches}
          \label{2nd_panda_image}
    \end{subfigure}
    \begin{subfigure}{0.33\textwidth}
          \includegraphics[width=\linewidth]{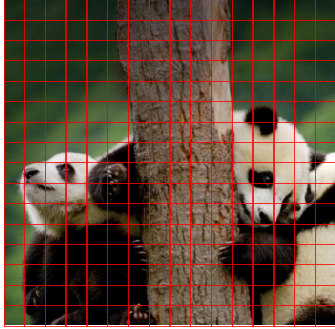}
          \caption{Lowest Level Patches}
          \label{lowest_panda_image}
    \end{subfigure}\hfil 
    \caption{Grid projection on an image of a panda sitting behind a tree.}
    \label{panda_image}
\end{figure}

Another benefit of creating multiple levels of patches versus only a single grid is that they would enable model to converge to an understanding about the whole image in fewer graphs iterations, this is explained by the following example. As shown in Figure \ref{panda_image}, we have an image of panda sitting behind a tree. If we only took the patches from a single grid (Figure \ref{lowest_panda_image}), then there would be patches on the left and right of the tree containing parts of a panda. These patches would query the middle part of the image and expect the presence of a panda's body. However, they'll find these patches to be quite dissimilar to their initial assumption (since a tree is in front of the panda). As there is no higher level node which can explain this absence of the panda in middle region, it'll take a larger number of graph iterations to reach the conclusion that there is a tree in front of the panda. However as shown in Figure \ref{highest_panda_image} and \ref{2nd_panda_image}, if the model has access to multiple levels of patches, once the assumptions of patches of lower levels are found out to be in-correct, they can directly query the higher level patches to provide an explanation for this. This direct source of information would help the model to reach the conclusion about the panda being behind a tree in fewer number of graph iterations. 

In short, during the feature extraction phase, patches in the higher levels learn to incorporate more information about the global structure of the image (i.e., large spatial context) while those in the lower levels focus more on understanding the local structure (i.e., fine details). During the graph iteration phase, the lower level patches can send queries to higher level patches to resolve conflicts among themselves and the high level patches can send queries to low level patch to confirm their assumptions and receive more information about contents at particular locations.

The issue about selecting which model (feature extractor) would work on which scale can be resolved simply by applying all the models on all these patches (and hence on all scales). A secondary model is then trained to determine which of the representations provided by the set of models actually represents the patch. Instead of doing a hard selection, we implement a soft selection method where we first determine the probability that a given feature extractor represents the patch and then do the weighted sum of all the feature representations using these probabilities.

There is however one drawback of dividing the image into patches using multiple levels of static grids (Figure \ref{static_generator}). This method assumes that information is uniformly distributed throughout the image, while in reality many of the patches generated would only contain the background of the image. Most images only contain a few objects of importance, humans can easily distinguish these objects from the background of the image and then focus on the portion of the image that contains the most information. We can design dynamic grid generation techniques (Figure \ref{dynamic_generator}) such that regions of high information content are covered in-depth (i.e., with both higher and lower level patches), while areas of low information like backgrounds are only explored briefly (i.e., with only higher level patches). This dynamic generation of patches would allow the model to use its resources in an efficient manner. Currently the idea is to create a heuristic for information content of a portion of image by utilizing simple image statistics. This can be later replaced by handcrafted models using traditional computer vision or neural networks.

\section{Model Architecture}
\begin{figure}[t]
    \centering
    \includegraphics[width=0.7\linewidth,keepaspectratio]{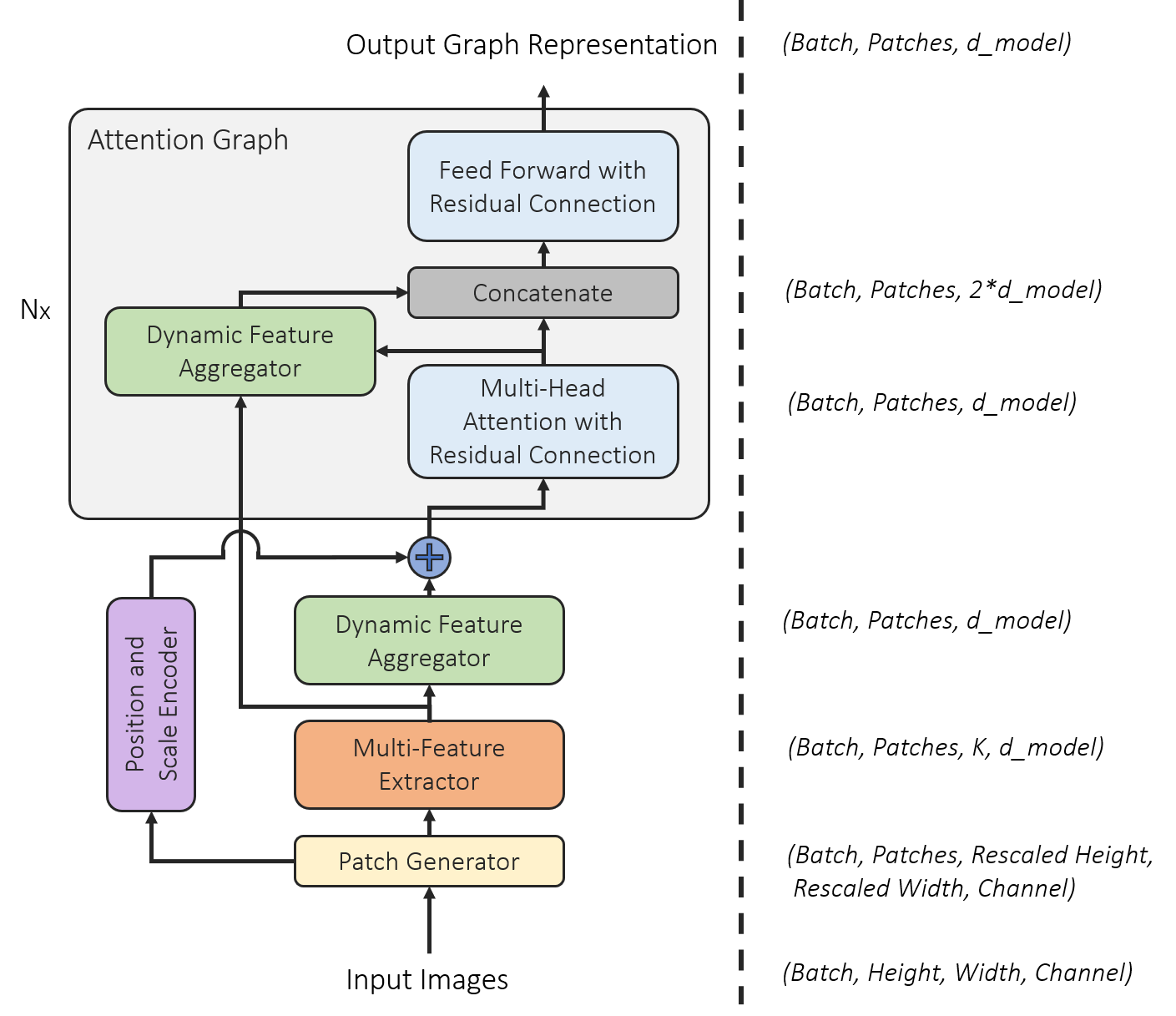}
    \caption{HindSight - Model Architecture.}
    \label{model_architecture}
\end{figure}

The model architecture consists of 5 main modules as shown in Figure \ref{model_architecture}.
\setlist{nolistsep}
\begin{enumerate}[noitemsep]
    \item \nameref{patch_generator}
    \item \nameref{feature_extractor}
    \item \nameref{feature_aggregator}
    \item \nameref{pos_scale_encoder}
    \item \nameref{attention_graph}
\end{enumerate}

Patch Generator takes the image as input and breaks the image into patches of different levels. All patches are re-scaled to be of same dimensions. Re-scaled patches are then processed by the multi-feature extractor, which returns multiple feature representations for each individual patch. Dynamic feature aggregator combines these multiple feature representations into a single feature vector per patch. Information about position (x and y coordinates) and area coverage of each patch is then added to form the final feature vector.

The final feature vectors are used as nodes for a fully connected graph using MHA (Multi-Head Attention) architecture for message passing between the nodes. Each layer of the MHA block along with a feedforward network represents one iteration of Graph Neural Network. Hence the number of attention graphs blocks determine the number of iterations of the graph network. Output of MHA layer can also be used as a query to dynamic feature aggregator. This allows the attention graph to verify its current understanding about the patch as well as extract extra information if needed. The result from dynamic feature aggregator is combined with the output of MHA layer and passed through the feedforward network to get the updated node representations for the given image.  

Hence each attention graph layer refines and updates the graph representation of the original image. The final graph representation can be used for various vision related tasks like Image Classification, Object Detection, Set Detection, Image Captioning, Visual Question Answering, Scene Graph Generation etc. 

The most important variables of the model architecture are $\bm{k}$ (number of patch levels and feature extractors), $\bm{N}$ (number of attention graph layers), $\bm{d}_{model}$ (size of latent dimension) and $\bm{H}$ (size of re-scaled image patch).

\subsection{Patch Generator}
\label{patch_generator}

\begin{figure}[t]
    \centering 
    \begin{subfigure}{0.33\textwidth}
          \includegraphics[width=\linewidth]{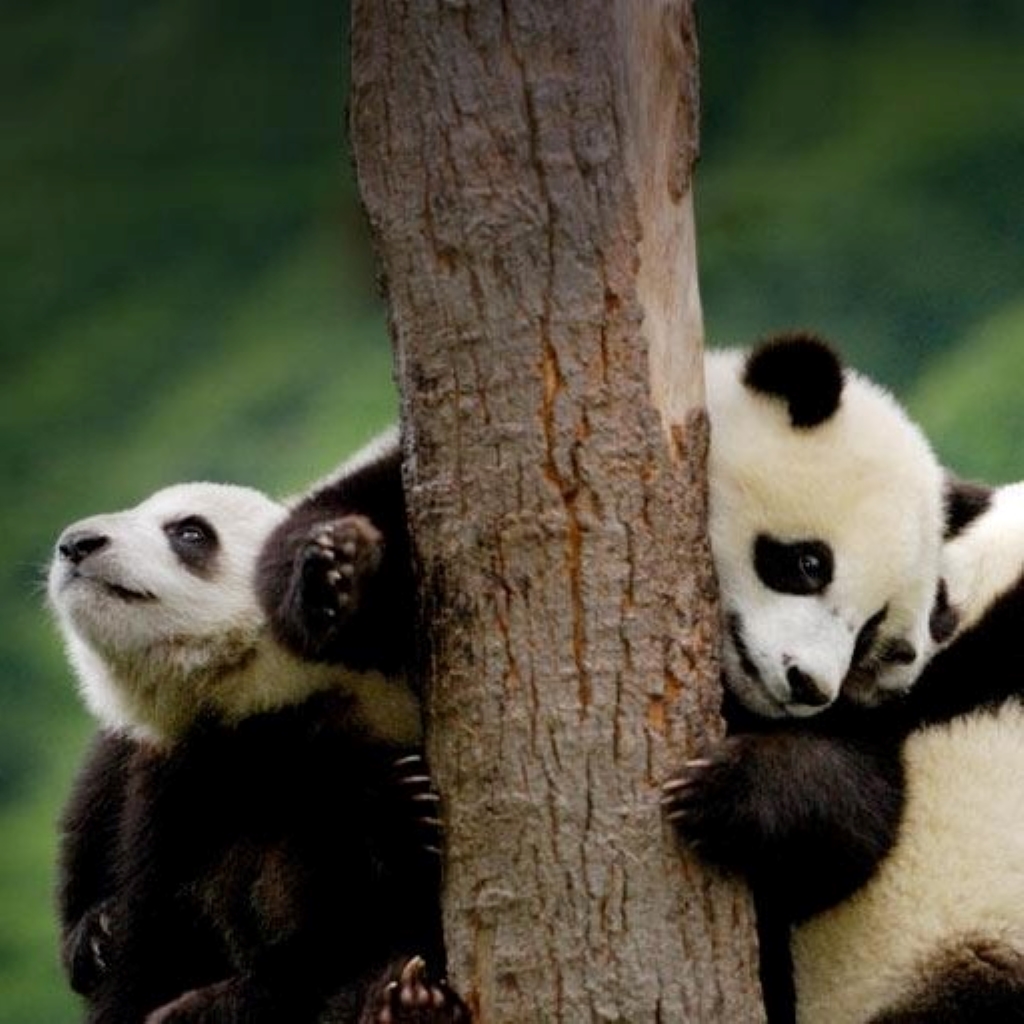}
          \caption{Original Image}
    \end{subfigure}\hfil 
    \begin{subfigure}{0.33\textwidth}
          \includegraphics[width=\linewidth]{1.png}
          \caption{Level 1: 1 Patch}
    \end{subfigure}\hfil 
    \begin{subfigure}{0.33\textwidth}
          \includegraphics[width=\linewidth]{2.png}
          \caption{Level 2: 4 Patches}
    \end{subfigure}
    
    \medskip
    \begin{subfigure}{0.33\textwidth}
          \includegraphics[width=\linewidth]{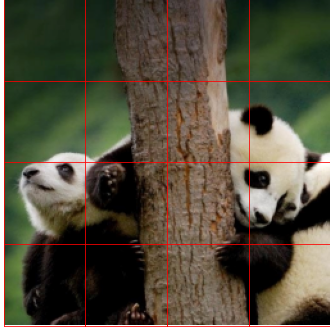}
          \caption{Level 3: 16 Patches}
    \end{subfigure}\hfil 
    \begin{subfigure}{0.33\textwidth}
          \includegraphics[width=\linewidth]{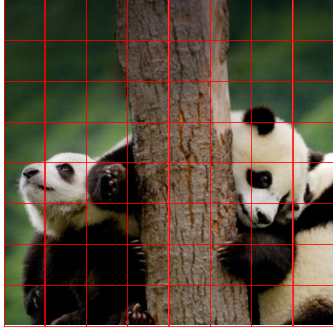}
          \caption{Level 4: 64 Patches}
    \end{subfigure}\hfil 
    \begin{subfigure}{0.33\textwidth}
          \includegraphics[width=\linewidth]{16.png}
          \caption{Level 5: 256 Patches}
    \end{subfigure}
    \caption{An example of patch generation using static grid with \textit{k=5} and input dimension of 1024x1024. All patches have been re-scaled to 64x64.}
    \label{static_generator}
\end{figure}

The purpose of this module is to divide the image into patches of multiple levels. 

This module is an input pre-processing step, which is currently proposed as non-parametric and deterministic. The most straightforward technique for the module (i.e., Static Regular Grids) is explained in this section. Dynamic Grid Generation techniques are discussed in Section \ref{intuition_continued}.

\paragraph{Static Regular Grids} This simple technique can be applied to square images with input dimensions $I$x$I$. The patch generation process consists of simply projecting a grid of $\{1,4,16,...,4^{k-1}\}$ squares for each level $k$ and then re-scaling all the resulting patches into same dimension of $H$x$H$. For example in Figure \ref{static_generator}, given an input image of 1024x1024 dimensions, the image is divided into 5 levels with $\{1,4,16,64,256\}$ patches each. The resulting patches can be re-scaled into any pre-selected dimension (in accordance to \nameref{feature_extractor}). In this example, $H=64$ and hence all the resulting patches are re-scaled to 64x64 (where 64x64 is also the $k^{th}$ level patch dimension). All of these patches are then concatenated to get the first output of this module. The overall number of patches $(P)$ grow exponentially with number of levels. Hence the first output of the module is defined by vector $X'$ which has length $P$.
$$P = \sum_{i=1}^{k} 4^{i-1} $$

For obtaining the second output of this module, a 2d co-ordinate system ($x,y$) is assigned to the image. This co-ordinate system assumes the value of $(0,0)$ at the center of the image. $x$ and $y$ can take values in the range of $[-1,1]$. In case the original image is not a square, the range for longer axis is kept as $[-1,1]$ while the range for shorter axis is reduced to preserve aspect ratio. Another variable $A$ ($Area_{Coverage}$) is also defined for each patch as following:

$$ Area_{Ratio} = \frac{Area_{Patch}}{Area_{Complete Image}} $$
$$ Area_{Coverage} =  1 + \frac{2 \log_{4}(Area_{Ratio})}{k} $$

This ensures that the value of $A$ stays within the range $[-1,1]$.The second output of the module is defined by vector $M \{(x_{1},y_{1},A_{1}),(x_{2},y_{2},A_{2}),...,(x_{P},y_{P},A_{P})\}$. Where each value corresponds to a node/patch in vector $X'$ on the same index.

\subsection{Multi-Feature Extractor}
\label{feature_extractor}

The purpose of this module is to generate \textit{k} many feature vectors from each given patch of image. 

\textit{k} many architecturally identical sub-modules are created with independent parameter weights. The output of this module is simply a concatenation of the outputs of all these sub-modules and is denoted as Multi-Feature Vector (\textit{MFV}). \textit{MFV} has dimensions of (\textit{k}, $d_{model}$).

\paragraph{Sub-Module} The key property that should be present in the sub-module is translation in-variance. Hence, the suggested backbone to achieve this is a CNN. Ideally the CNN's \textit{Conv + Pool} layers should reduce the dimension of the patch to (1x1) before flattening. For example, for $H$ being 64 and $d_{model}$ being \textit{512} the CNN structure shown in Table \ref{sub_model_structure} can be implemented. This would then be flattened and followed by 1-2 dense layers of $d_{model}$ dimensions to obtain the output feature vector (\textit{FV}) of the respective sub-module.

\begin{table}[H]
\captionof{table}{Sample Structure for CNN Sub-Module.}\vspace{8pt}
 \label{sub_model_structure} 
\centering
\begin{tabular}{lcc}
\toprule
\textbf{Layer} & \textbf{Dimensions} & \textbf{Filters/Channels} \\ \hline
Input         & 64 x 64    & 3                \\
Conv (5,5)    & 60 x 60    & 32               \\
Conv (5,5)    & 56 x 56    & 32               \\
MaxPool (2,2) & 28 x 28    & 32               \\
Conv (3,3)    & 26 x 26    & 64               \\
Conv (3,3)    & 24 x 24    & 64              \\
MaxPool (2,2) & 12 x 12    & 64              \\
Conv (3,3)    & 10 x 10    & 128              \\
Conv (3,3)    & 8 x 8      & 128              \\
Conv (3,3)    & 6 x 6      & 128              \\
MaxPool (2,2) & 3 x 3      & 128              \\
Conv (3,3)    & 1 x 1      & 256             \\
\bottomrule
\end{tabular}
\vspace{5pt}
\end{table}

The idea is that each sub-module would learn a different representation of the patch. Let $FV^{(j)} \in R^{d_{model}}$ denote the $j$-th Feature Vector in Multiple Feature Vector (\textit{MFV}). Such that for each patch:
$$ MFV = [FV^{(1)},FV^{(2)},...,FV^{(k)}] $$

\subsection{Dynamic Feature Aggregator}
\label{feature_aggregator}

The purpose of this module is to provide the model with a mechanism to query the multiple feature representations and receive an Aggregated Feature Vector (\textit{AFV}) for each patch. 

The module takes the \textit{MFV} as input by default, an additional input of Graph-Query ($GQ \in R^{d_{model}}$) is provided inside the \nameref{attention_graph} Module. If \textit{GQ} is not provided, it is initialized as a zero vector ($\bm0^{d_{model}}$). The whole module can be described as:

$$ \eta_{gq} = ReLu(\bm{W}_{1}GQ + \bm{b}_{1})$$
$$ \eta_{1}^{(j)} =  ReLu(\bm{W}_{2,fv} FV^{(j)} + \bm{W}_{2,gq}\eta_{gq}  + \bm{b}_{2})$$
$$ \eta_{2}^{(j)} =  \bm{W}_{3} \eta_{1}^{(j)} + \bm{b}_{3}$$
$$ C^{(j)} = \frac{\exp(\eta_{2}^{(j)})}{\sum_{i=1}^{k}\exp(\eta_{2}^{(i)})} $$
$$ AFV = \sum_{j=1}^{k} C^{(j)} FV^{(j)} $$

All \textit{k} feature vectors in \textit{MFV} (from \nameref{feature_extractor}) along with \textit{GQ} (from \nameref{attention_graph}) are passed through the same feed-forward network. This results in a scalar value ($\eta_{2}$) corresponding to each provided \textit{FV}. A Softmax function is used to normalize these values and finally a weighted sum is performed over the \textit{MFV} using these normalized values (\textit{C}).

Since the feed-forward network shares its parameters across all the feature vectors, we can introduce the negative Kullback–Leibler divergence, of the distribution of $\bm{C}$ vector from a uniform distribution of \textit{k} elements, as another loss during training. Addition of this loss would ensure that each of the sub-module learns a different representation from the same image patch. This loss would only be applied to the Dynamic Feature Extractor module outside the Attention Graph module. 

$$ L_{divergent} =  -I\mkern-8muD_{{KL}}(\bm{C}\|Uniform Dist(k))$$

The idea is that each feature extractor is forced to learn a different aspect from the image patch. Initially, the FV are aggregated only using the information from the patch. However, the further modules of the model can query the Dynamic Feature Aggregator for specific information via the graph-query ($GQ$) vector. 

Currently all layers of the graph network share the same $\bm{W}_{1}$ and $\bm{b}_{1}$ parameters, hence they are invariant of the graph depth. Alternatively, $\eta_{gq}$ can also be defined inside \nameref{attention_graph} module instead. This would enable graph network at each depth to send in a customized query.

\subsection{Position and Scale Encoder}
\label{pos_scale_encoder}

The purpose of this module is to provide each patch a sense of its location and coverage with respect to the overall image. 

In this module, the vector \textit{M} is first transformed into a Encoding Vector (\textit{EV}) and then added to \textit{AFV}, the result is denoted by Final Feature Vector (\textit{FFV}). 

$$ FFV_{node} = AFV_{node} + EV_{node} $$

Some general methods are discussed below:

\paragraph{Trainable Encoding} The most simple way to encode this information is to use a small feed-forward network of $d_{model}$ dimensions. Here, we learn a continuous embedding function from variable space to \textit{AFV} space. Due to only a few number of unique values of each variable, this method may find it difficult to inter/extrapolate for previously unseen values using testing.

\paragraph{Non-Trainable Periodic Encoding} Another popular encoding technique which is implemented using sine and cosine functions is also applicable here. Because of its continuous nature, it should allow the model to better interpolate the unseen values encountered during testing. It is suggested that x, y and area variables are encoded in $d_{model}/4$, $d_{model}/4$ and $d_{model}/2$ dimensions respectively. Here $\bm{EV}$ is given by concatenating $[EV^{x},EV^{y},EV^{area}]$. Where \textit{EV} for each variable (\textit{var}) is defined as:

$$ EV^{^{var}}_{(node,2i)} = sin(var_{node}\lambda^{i/d_{model}} )  $$
$$ EV^{^{var}}_{(node,2i+1)} = cos(var_{node}\lambda^{i/d_{model}} )  $$

Where \textit{i} ranges from \textit{0} to \textit{(pre-determined encoding dimension)/2} for that given \textit{var} and $\lambda$ is a hyper-parameter. The ideal value for $\lambda$ would be found empirically.

\paragraph{Trainable Periodic Encoding} One concern of using non-trainable periodic encoding is that it forces both the \textit{AFV} and \textit{EV} to co-habit the same vector space. Since \textit{EV} is defined by non-trainable parameters, hence only \textit{AFV} is actually affected. This may limit the expressive power of the \textit{AFV}. A simple way to remedy this is to first construct an intermediate encoding vector by using the non-trainable periodic encoding method and then pass the it through a small feed-forward network. The resultant vector is then added to \textit{AFV}. In this method, the feed-forward network should project the periodic encoded vector into aggregated feature vector space and hence introduce less disturbance/noise as compared to non-trainable periodic encoding method while preserving its inter/extrapolating property. $\lambda$ can simply be set to $1$ in this setting.

\subsection{Attention Graph}
\label{attention_graph}

The purpose of this module is to explore the relationship between patches (both inter-level and intra-level) and refine the overall graph representation of the image.

The module can be constructed using any GNN mechanism, however recent research (\cite{VitPaper}, \cite{E2E_OD}, \cite{DEIT}) has demonstrated great performance of transformers for vision tasks. Hence, the transformer encoder module is proposed for this task with a small modification as shown in Figure \ref{model_architecture}. The inputs (\textit{FFVs}) are used as tokens for the Multi-Head Attention module, the outputs from the MHA (after residual connection and layernorm) are then used as Graph-Query (\textit{GQ}) to the Dynamic Feature Aggregator. The returned \textit{FV} is concatenated to the MHA output and passed through the feedforward network. This small modification allows model the to explore and utilize the multiplicity of \nameref{feature_extractor}. 

In order to reduce some model complexity, \nameref{feature_aggregator} can be incorporated only in every \textit{i}-th layer of the attention graph.

\section{Intuition Continued}
\label{intuition_continued}

\paragraph{Dynamic Grids} All downstream computation is dependent on the initial technique used for splitting the image into multiple levels of patches. Ideally each patch of the image should either compass an object or a part of the object. Section \ref{feature_extractor} described the workings of Static Grid Generator. The drawback of using static grid generator is that all regions of the image are divided into square patches and then explored with same number of levels. This is an in-efficient use of resources (computation). We can improve the resource allocation by using dynamic grid generation techniques, where we focus more on regions of image that have a higher information content. Since the actual information content of the image is not known, we need to develop some heuristics as a substitute.

\begin{figure}[b!]
    \begin{subfigure}{0.3\textwidth}
          \includegraphics[width=\linewidth]{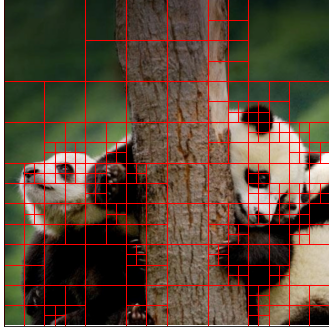}
          \caption{Dynamic Grid Projection}
    \end{subfigure}\hfil 
    \begin{subfigure}{0.3\textwidth}
          \includegraphics[width=\linewidth]{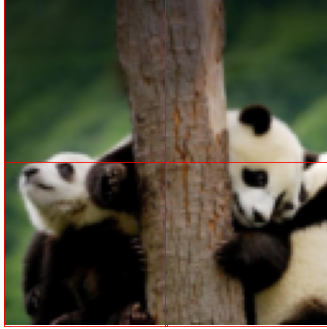}
          \caption{Level 2: 4 Patches}
    \end{subfigure}\hfil 
    \begin{subfigure}{0.3\textwidth}
          \includegraphics[width=\linewidth]{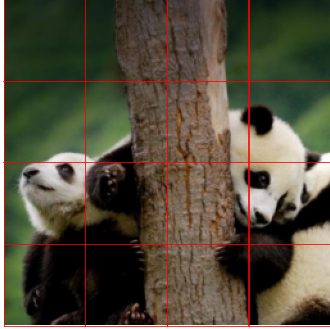}
          \caption{Level 3: 16 Patches}
    \end{subfigure}
    
    \begin{subfigure}{0.3\textwidth}
          \includegraphics[width=\linewidth]{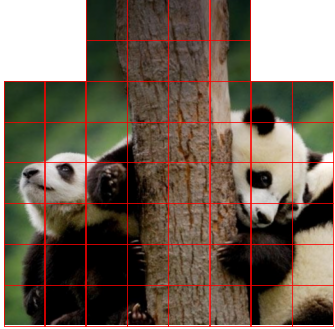}
          \caption{Level 4: 56 Patches}
    \end{subfigure}\hfil 
    \begin{subfigure}{0.3\textwidth}
          \includegraphics[width=\linewidth]{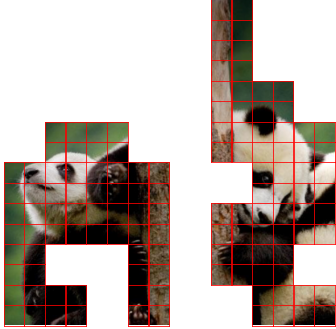}
          \caption{Level 5: 124 Patches}
    \end{subfigure}\hfil 
    \begin{subfigure}{0.3\textwidth}
          \includegraphics[width=\linewidth]{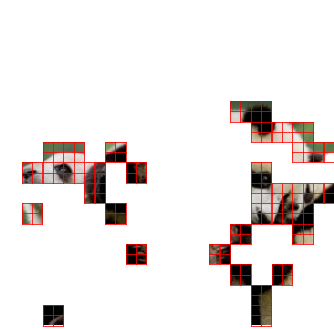}
          \caption{Level 6: 140 Patches}
    \end{subfigure}
    \caption{An example of patch generation using dynamic grid with $k = 6$, $D = 85$ and $P = 341$. All patches have been re-scaled to 64x64.}
    \label{dynamic_generator}
\end{figure}

The simplest method to create a dynamic grid is to construct a Quadtree from the image where we divide the node with the highest information content (as defined by a heuristic) and repeat the process till the desired number of patches is reached. Figure \ref{dynamic_generator} is an examples of grid generation using this method, here the number of patches is equal to number of patches in the static grid generator (Figure \ref{static_generator}) however we have allowed the model to explore one more level (i.e., $k=6$).

This method allows us to explore more levels without the exponential growth in number of patches. In this example, the heuristic used is the mean squared deviation from average colour of the patch. This heuristic was inspired from a Quadtree Visualization (\url{https://www.michaelfogleman.com/static/quads/}). Two parameters required by the grid generation algorithm include $k$ (number of levels) and $D$ (number of patches to divide). $P$ (number of overall patches) is related to $D$ by the following simple equation:
$$P = 1 + 4D$$

The algorithm for dynamic grid generation is quite straightforward. Construct a Quadtree with whole image as the first node and divide the node into 4 child nodes by dividing the image patch into 4 equal patches. Calculate the information content of these new child nodes. Find the child node with highest information content and repeat the process $D$ times. We also check the level of node in the tree and only divide the node if the level is less than $k$. The resulting quadtree represents our dynamic grid.

\paragraph{Grids \& Part-Whole Hierarchies} \label{peusdo_part_whole}

This systematic creation of a grid and division of image into patches is done to provide a coarse sense of hierarchy to the model. Some may argue that the boundaries of objects and their parts are mostly irregular in nature. Therefore, the true part-whole hierarchies could only be learned if the region containing each object/part is defined by exact bounding boxes/pixel segmentation for each of the respective object/part. Since this method only divides the image up into square patches (where each patch would contain \textit{(A)} a complete object, \textit{(B)} a partial object, \textit{(C)} a complete part or \textit{(D)} a partial part), hence it would be in-sufficient to learn this property. It is wise to remember that achieving the representation of images which encompass the information about part-whole hierarchies is the end goal of the model and not its input requirement.

The model contains a fully connected graph where each node is connected to all others instead of just the parent patch being connected to the child patches. This enables the model to revise all the node representations and refine the relationship between the nodes. A querying mechanism is also provided to the model to re-affirm its assumption directly from the image patches. So even if the original node representation of a patch may not exactly encompass the true representation of its contents and the complete context, this information (if it exists in the image) could always be gathered by querying the feature extractors or by interacting with other higher, same or lower level patches in the graph. 

In the resulting refined graph, each node would contain the information about its own content, how is it related to the overall global structure and what finer details can be found within it.

\section{Model Size and Training}
\label{training}

\paragraph{Model Size} Size of a model built using HindSight architecture can be defined by $\bm{k}$ (number of patch levels and feature extractors), $\bm{N}$ (number of attention graph layers), $\bm{d}_{model}$ (size of latent dimension) and $\bm{H}$ (size of re-scaled patch). It is suggested that $\bm{k}$ should be between 4-6. For larger images (greater than 1 MegaPixel) dynamic grids could be implemented with a limit on the number of patches ($P$) and higher $\bm{k}$ levels (5-7). The patches should be re-scaled to dimension ($\bm{H}$) between 32-128. If the patches are too small, they hinder the higher level patches to be meaningful and if the patches are too large, the CNN's structural priors start affecting quality of feature representations being generated.

\paragraph{Training} The most straightforward method to train the model would be to use a supervised learning task of Image Classification using a sufficiently large dataset (like JFT-300M), which would require a large amount of computational resources. However, this approach does not fully utilize the learning capacity that HindSight provides us. A model trained in a supervised learning task like Image Classification would only learn the distinct features of each class that enables the model to distinguish this class from other input classes. Apart from these distinct features, the remaining information present in the image is not utilized by the model at all. Size of dataset used in supervised training is also limited since it requires careful labelling by humans. Hence, a better approach to train HindSight would be to implement it as an image encoder for a pretext task in a self-supervised setting. \cite{inpainting} (Context Encoders) provides an ideal starting point as the pretext task. However, some refinements may be required in the self-supervised training techniques to enable the model to deal with the multi-modal nature of images.

\begin{figure}[t]
    \centering
    \begin{subfigure}{0.3\textwidth}
          \includegraphics[width=\linewidth]{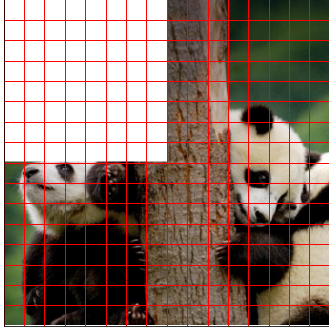}
          \caption{Masking at Level 2}
    \end{subfigure}
    \hfil 
    \begin{subfigure}{0.3\textwidth}
          \includegraphics[width=\linewidth]{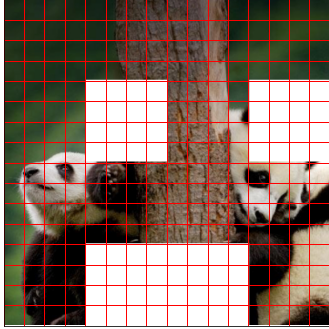}
          \caption{Masking at Level 3}
          \label{mask_l3}
    \end{subfigure}

    \begin{subfigure}{0.3\textwidth}
          \includegraphics[width=\linewidth]{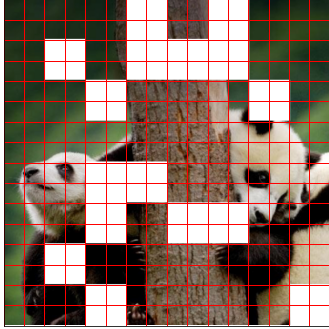}
          \caption{Masking at Level 4}
    \end{subfigure}
    \hfil 
    \begin{subfigure}{0.3\textwidth}
          \includegraphics[width=\linewidth]{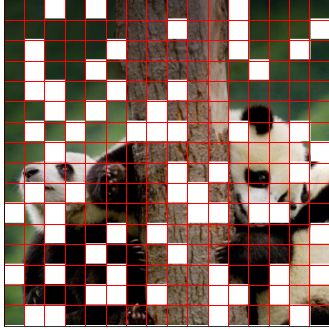}
          \caption{Masking at Level 5}
    \end{subfigure}
    \caption{An example of mask generations for different grid levels. All of these samples represent different training samples for the un-supervised task}
    \label{mask_generator}
\end{figure}

During self-supervised training, the model would have to predict contents of masked out regions in an image. As shown in Figure \ref{mask_generator}, it is first suggested to select uniformly at random any level $k>1$ and then mask out the regions that $1/4$ (or $1/8$) of the patches in this level covers. As masking is performed before the patch generator, higher level patches would contain some portions of masked regions while some lower level patches would only contain masked regions. This variation in masking at different levels would encourage the model to learn the part-whole hierarchies of objects and the inherent structural relationship between them. For the decoder part of the pretext task, it is suggested to use a transformer model which takes as input the positional and area encoding vector $M$ corresponding to all the patches which have been completely masked. The model is trained to provide a pixel-level reconstruction for each of these patches. For example, if $k=3$ (Figure \ref{mask_l3}) then we would have a total of 84 patches for the decoder to reconstruct (4,16 and 64 patches from level 3,4 and 5 respectively).

Training in this self-supervised manner would require even more data and computational resources than supervised training, however once trained the resultant encoder model would be more generalized and would only require some fine-tuning when adapting to various downstream tasks.

\section{Discussion}

\paragraph{Self-Critique} Machine learning is a highly empirical discipline, one of the most import aspect of an architecture is how well it performs versus the current state of the art models. Recently introduced vision models (which do not rely on CNNs as the main module), usually require large datasets and huge compute resources for their training. The amount of data and computation resources needed to train HindSight (Section \ref{training}) as a generalized vision model are a little beyond the reach of a single individual. 

\paragraph{Direction of Research} While this paper was being written, two papers (\citet{wu2021cvt} and \citet{chen2021crossvit}) were published which explored some modifications in the Vision Transformer (\citet{VitPaper}) architecture and achieved better performances.

CvT (\citet{wu2021cvt}) combines convolutions with transformers such that it utilizes a convolutional embedding block and a convolutional transformer block in each stage of the model. CvT implements three such stages, where in each stage the model takes the finer details from the lower level of patches and then convolves them in representation space to get the representations of the higher levels. Eventually at the end, only higher level representations are left which are then utilized to perform the task on hand (Image Classification). CvT never allows the higher and lower level patches to interact together in the transformer blocks, hence the global structure of image is only learned by combining the outputs of previous layers. The model does not have any direct connection to the original source of information for the global structure and is thus highly dependent on the encoding done at lower levels. 

CrossViT (\citet{chen2021crossvit}) generates the image patches using two different grid levels. Compared to HindSight, the difference in scale of these two levels is every small and both of these different set of patches go through different feature extraction modules. Furthermore, CrossVit does not include higher level patches that would contain the global context of the image and only rely on two lower level patches (like ViT and CvT). Recent research which explores such similar ideas reassures me that HindSight is indeed a step in the right direction.

\paragraph{Conclusion} All concepts, methods, techniques and modules discussed in this paper have already been researched and implemented successfully in different architectures. HindSight is just a unique amalgamation of these ideas which would enable us to represent part-whole hierarchies in images in form of a graph. If HindSight can achieve this, then it has the potential to be a general purpose vision encoder model (similar to current language models in NLP). However, it still may require a few more insights and refinements to reach this stage. The reason of sharing this model architecture (without even a working model) is to engage with researchers and collaborators that can help in finding these few insights and refinements. 

\setcitestyle{numbers}
\bibliographystyle{unsrtnat}
\bibliography{references}

\end{document}